\documentclass[journal]{IEEEtran}
\usepackage{epstopdf}
\usepackage{tabularx,threeparttable}
\usepackage{array}
\usepackage{hyperref}

\usepackage{graphicx}
\usepackage{subfigure}
\usepackage{amssymb, amsmath, bm}
\usepackage{mathtools}
\usepackage{mathrsfs}
\usepackage{booktabs}
\usepackage{multirow}
\usepackage{cite}
\usepackage{url}
\usepackage{color}
\usepackage{dsfont}

\def\eg{\textit{e.g.}}
\def\ie{\textit{i.e.}}
\def\etal{\textit{et al.}}

\newcommand\T{\rule{0pt}{2.6ex}}       
\newcommand\B{\rule[-1.2ex]{0pt}{0pt}} 
\newcolumntype{?}[1]{!{\vrule width #1}}

\ifCLASSINFOpdf
\else
\fi
\begin{document}
%
\title{Patch-based Output Space Adversarial Learning \\ for Joint Optic Disc and Cup Segmentation}
%
%
%

\author{Shujun Wang,
	Lequan Yu,~\IEEEmembership{Student~Member,~IEEE,}
	Xin Yang,
	Chi-Wing Fu,~\IEEEmembership{Member,~IEEE,} \\
	and Pheng-Ann Heng,~\IEEEmembership{Senior~Member,~IEEE}
	
\thanks{S. Wang, L. Yu,  X. Yang are with the Department of Computer Science and Engineering, The Chinese University of Hong Kong, Hong Kong (e-mail: sjwang@cse.cuhk.edu.hk; lqyu@cse.cuhk.edu.hk; xinyang@cse.cuhk.edu.hk).}
\thanks{C.-W. Fu and P.-A. Heng are with the Department of Computer Science and Engineering, The Chinese University of Hong Kong, Hong Kong, and also with the Guangdong Provincial Key Laboratory of Computer Vision and Virtual Reality Technology, Shenzhen Institutes of Advanced Technology, Chinese Academy of Sciences, Shenzhen, China (e-mail: cwfu@cse.cuhk.edu.hk; pheng@cse.cuhk.edu.hk).}
\thanks{C.-W. Fu and P.-A. Heng are the co-corresponding authors of this work.}
\thanks{Copyright (c) 2018 IEEE. Personal use of this material is permitted. Permission from IEEE must be obtained for all other uses, including reprinting/republishing this material for advertising or promotional purposes, collecting new collected works for resale or redistribution to servers or lists, or reuse of any copyrighted component of this work in other works. Citation information: DOI 10.1109/TMI.2019.2899910.}
}

\maketitle

%

\begin{abstract}
	Glaucoma is a leading cause of irreversible blindness.
	Accurate segmentation of the optic disc (OD) and cup (OC) from fundus images is beneficial to glaucoma screening and diagnosis.
	Recently, convolutional neural networks demonstrate promising progress in joint OD and OC segmentation.
	However, affected by the domain shift among different datasets, deep networks are severely hindered in generalizing across different scanners and institutions.
	In this paper, we present a novel patch-based Output Space Adversarial Learning framework (\textit{p}OSAL) to jointly and robustly segment the OD and OC from different fundus image datasets.
	We first devise a lightweight and efficient segmentation network as a backbone. Considering the specific morphology of OD and OC, a novel morphology-aware segmentation loss is proposed to guide the network to generate accurate and smooth segmentation.
	Our \textit{p}OSAL framework then exploits unsupervised domain adaptation to address the domain shift challenge by encouraging the segmentation in the target domain to be similar to the source ones. Since the whole-segmentation-based adversarial loss is insufficient to drive the network to capture segmentation details, we further design the \textit{p}OSAL in a patch-based fashion to enable fine-grained discrimination on local segmentation details.
	We extensively evaluate our \textit{p}OSAL framework and demonstrate its effectiveness in improving the segmentation performance on three public retinal fundus image datasets, \ie, Drishti-GS, RIM-ONE-r3, and REFUGE.
	Furthermore, our \textit{p}OSAL framework achieved the first place in the OD and OC segmentation tasks in \textit{MICCAI 2018 Retinal Fundus Glaucoma Challenge}.

\end{abstract}


\section{Introduction}

\begin{figure}
	\centering
	\includegraphics[width=0.48\textwidth]{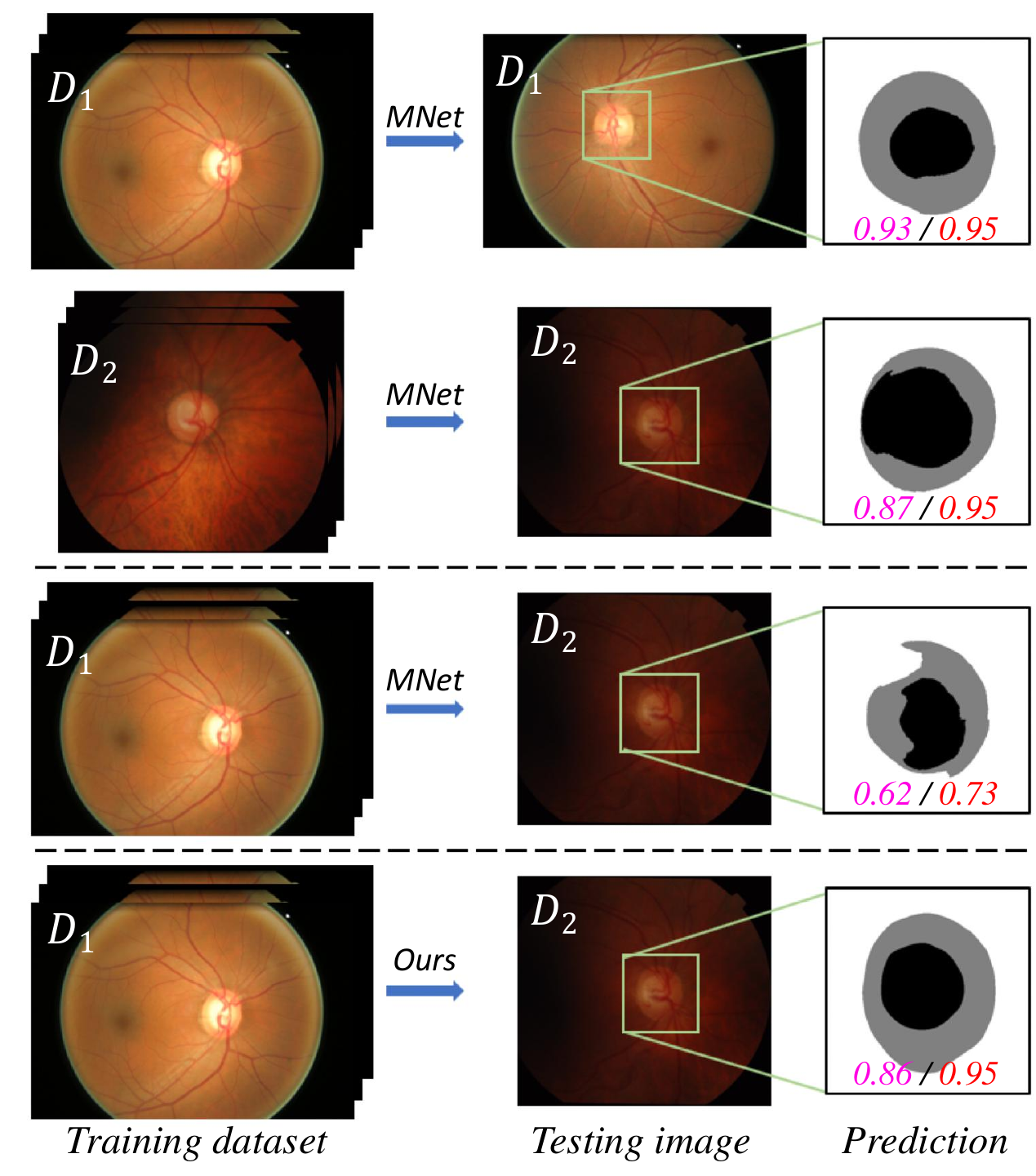}
	\caption{Segmentation degradation due to domain shift. Domain $D_1$ stands for the ORIGA dataset, while domain $D_2$ stands for the Drishti-GS dataset. In the \textit{Prediction} column, the black and gray colors represent the optic cup (OC) and optic disc (OD) segmentation, respectively. The two numbers are the dice coefficients for the OC and OD segmentation results, showing that the dice coefficients degrade from 0.87 to 0.62 for OC and from 0.95 to 0.73 for OD, when we use the M-Net~\protect\cite{fu2018joint} trained on D1 to test on D2. Our method overcomes the problem; while it is trained on $D_1$, it still can achieve high dice coefficients of 0.86 and 0.95 for OC and OD, respectively, when using it on $D_2$.} 
	\label{fig:dice_drop}
	\vspace{-0.3cm}
\end{figure}

\IEEEPARstart{G}{laucoma} is a chronic disease that damages the optic nerves and leads to irreversible vision loss \cite{mary2016retinal}. 
Screening and detecting glaucoma in its early stage are beneficial to preserve the vision of patients. Currently, analyzing the optic nerve head and retinal nerve fiber layer is a practical method for glaucoma detection. However, such analysis is predominantly subjective and often suffers from the high intra- and inter-observer variations~\cite{naithani2007evaluation}.
With the recent advancement in optical fundus imaging, objective and quantitative glaucoma assessments based on the morphology of optic disc (OD) and optic cup (OC), and the cup-to-disc ratio (CDR) become available~\cite{mary2016retinal}.
CDR is the ratio of vertical cup diameter to vertical disc diameter. A large CDR value often indicates a high risk of glaucoma.
Manually acquiring those measurements is time-consuming. Accurately segmenting OD and OC from the fundus image via automatic solutions would prompt the large-scale glaucoma screening~\cite{fu2018joint}.

Remarkable performance on OD and OC segmentation are recently reported with the development of deep learning~\cite{maninis2016deep, shankaranarayana2017joint,fu2018joint}.
Assuming the training and testing samples have the same appearance distribution, training dataset consisting of a large amount of pixel-level annotations helps the deep networks learn the segmentation on the testing dataset.
However, it is difficult for the network to obtain good segmentation performance on new datasets. 
For example, the state-of-the-art network, like M-Net~\cite{fu2018joint} performs well on its specific testing dataset, \ie, ORIGA~\cite{zhang2010origa}, but generalizes poorly on some other datasets; see Fig.~\ref{fig:dice_drop}.
Domain shift, which refers to the difference in appearance distribution between the different datasets, is the main cause for the poor generalization ability of deep networks~\cite{ghafoorian2017transfer,kamnitsas2017unsupervised,dou2018unsupervised}.
Indeed, domain shift among various retinal fundus image datasets is very common.
Many public retinal image datasets, \eg, Drishti-GS~\cite{Drishti-GS}, RIM-ONE-r3~\cite{RIM-ONE}, and REFUGE, are acquired with obvious appearance discrepancy resulting from different scanners, image resolution ratios, light source intensities\textcolor{blue}{,} and parameter settings (Fig.~\ref{fig:dice_drop}). Overcoming the domain shift is highly desired to enhance the robustness of deep networks.

To reduce the performance degradation caused by domain shift, \textit{domain adaptation} methods~\cite{kamnitsas2017unsupervised,patel2015visual} are developed to generalize the deep networks trained in a source domain to work more effectively in some other target domains with varying appearance.
A vanilla solution is to fine-tune the segmentation network with a full-supervision provided by a large quantity of annotated samples from the target domain.
However, preparing for the extra annotations in the target domain is highly time-consuming and expensive, and often suffers from inter-observer variations; moreover, such a solution is impractical for large-scale glaucoma screening.
Therefore, an unsupervised domain adaptation approach without requiring extra annotations is highly desirable in real clinical scenarios.
Furthermore, leveraging the knowledge shared across different domains can help the deep networks maintain their performance under various imaging conditions.
For this joint OD and OC segmentation task, spatial and morphological structures in the output space (\ie, segmentation mask) are shared by different datasets and thus are beneficial to the mask prediction.
For example, the OC is always contained inside the OD region, while both the OC and OD have ellipse-like shapes. Such spatial correlation information is crucial for domain adaptation but is typically ignored by existing deep-network-based segmentation methods.

In this work, we aim at jointly segmenting the OD and OC in retinal fundus images from different domains by introducing a novel patch-based Output Space Adversarial Learning framework (\textit{p}OSAL). As the core workhorses in the framework, the lightweight network architecture for efficacy and the unsupervised domain adaptation for domain-invariance contribute to our promising performance.
%
Our framework explores the annotated source domain images and unannotated target domain images to reduce the performance degradation on target domain.
We first develop a representative segmentation network equipped with a morphology-aware segmentation loss to produce compelling segmentations. Effectively combining the designs of DeepLabv3+~\cite{chen2018encoder} and depth-wise separable convolutional network MobileNetV2 \cite{sandler2018mobilenetv2}, our segmentation network achieves a good balance between extracting multi-scale discriminative context features and computational burden.
The proposed morphology-aware segmentation loss further guides the network to capture mask smoothness priors and therefore improves the segmentation.
To overcome the domain shift challenge, inspired by~\cite{tsai2018learning}, we adopt the output space adversarial learning via utilizing the spatial and morphological structures of the segmentation mask.
Specifically, we attach a discriminator network to learn the abstract spatial and shape information from the label distributions of the source domain, and then employ the adversarial learning procedure to encourage the segmentation network to generate consistent predictions in a shared output space (\eg, the similar spatial layout and structure context) for the images in both source and target domains.
Since the whole-segmentation-based adversarial scheme is weak in capturing segmentation details,
we devise a patch-wise discriminator to capture the local statistics of the output space and guide the segmentation network to focus on the local structure similarity in the image patches.
We extensively evaluate our \textit{p}OSAL framework for the joint OD and OC segmentation on three public fundus image datasets (Drishti-GS, RIM-ONE-r3, and REFUGE).
The \textit{p}OSAL framework achieves state-of-the-art results, bringing significant improvements with the proposed patch-based output space adversarial learning.

Our main contributions are summarized as follows:
\begin{enumerate}
	\item
	We exploit unsupervised domain adaptation for joint OD and OC segmentation over different retinal fundus image datasets. The presented novel \textit{p}OSAL framework enables patch-based output space domain adaptation to reduce the segmentation performance degradation on target datasets with domain shift.
	\item 
	We design an efficient segmentation network equipped with a new morphology-aware segmentation loss to produce plausible OD and OC segmentation results. The morphological segmentation loss is able to guide the network to capture the mask smoothness priors for accurate segmentation.
	\item
	We conduct extensive experiments on three public retinal fundus image datasets to demonstrate the effectiveness of the \textit{p}OSAL framework. 
	Furthermore, we achieved the first place in the OD and OC segmentation task of the \textit{MICCAI 2018 Retinal Fundus Glaucoma Challenge}. 
	
\end{enumerate}

The remainders of this paper are organized as follows.
We review the related techniques in Section~\ref{sec:relatedwork} and elaborate the \textit{p}OSAL framework in Section~\ref{sec:method}.
The experiments and results are presented in Section~\ref{sec:experiment}.
We further discuss our method in Section~\ref{sec:discussion} and draw the conclusions in Section~\ref{sec:conclusion}.

\begin{figure*}
	\centering
	\includegraphics[width=\textwidth]{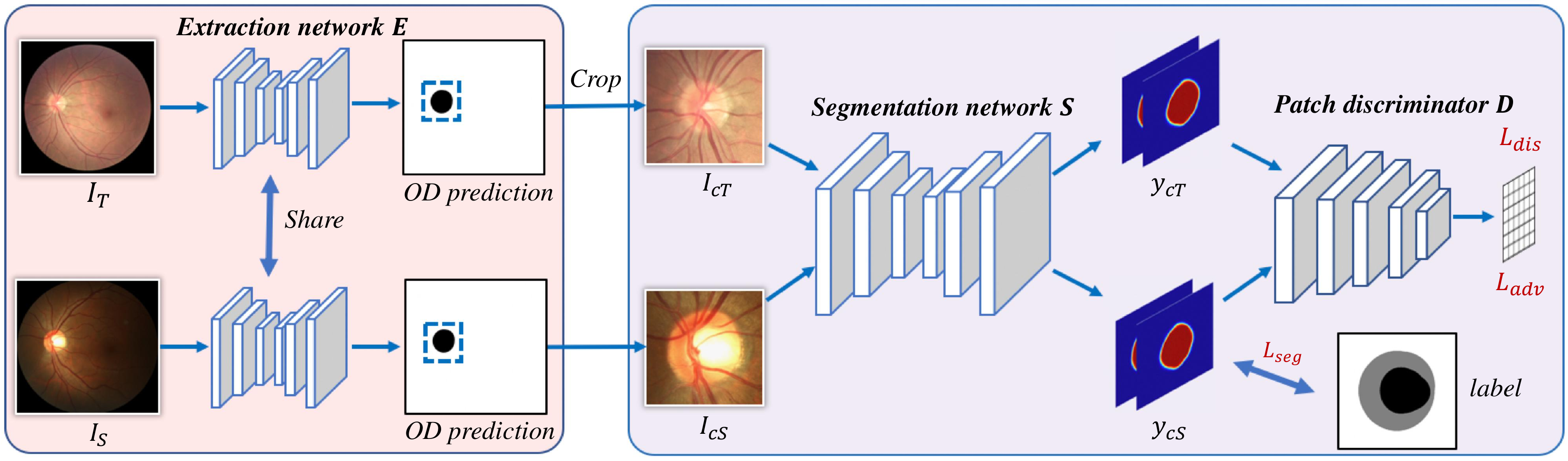}
	\caption{Overview of the \textit{p}OSAL framework. ROI regions ($I_{cT}, I_{cS}$) are firstly extracted from the source ($I_S$) and target ($I_T$) domain images and then fed into the segmentation network $S$. The discriminator $D$ in a patch-based adversarial learning scheme enforces the similarity between the target image prediction ($y_{cT}$) and source ones ($y_{cS}$). The segmentation network is supervised by the segmentation loss ($L_{seg}$) computed on the prediction of source domain images ($y_{cS}$) and the adversarial loss ($L_{adv}$) calculated on the prediction of unlabeled target domain images ($y_{cT}$).}
	\label{fig:overview}
\end{figure*}

\section{Related Works}
\label{sec:relatedwork}

The OD and OC segmentation from retinal fundus images are non-trivial and have been independently studied for years. For the OD segmentation, early works employed the hand-crafted visual features, including the image gradient information \cite{lowell2004optic}, features from stereo image pairs \cite{abramoff2007automated}, local texture features \cite{joshi2011optic} and superpixel-based classifier \cite{cheng2013superpixel}. The OC segmentation is more challenging than OD considering the lower-contrast boundary~\cite{fu2018joint}. Hand-crafted features are also investigated for this task~\cite{joshi2011optic,cheng2013superpixel,wong2008level,wong2009automated,xu2011sliding,xu2014optic}.
Recently, some works were developed for joint OD and OC segmentation. Zheng \etal~\cite{zheng2013optic} designed a graph-cut framework. In~\cite{xu2012efficient}, structure constraints were utilized for joint OD and OC segmentation.

Convolutional neural networks (CNNs) have shown remarkable performance on retinal fundus image segmentation~\cite{fu2018joint, maninis2016deep, shankaranarayana2017joint,zilly2017glaucoma,sevastopolsky2017optic, fu2018disc, edupuganti2018automatic}, and outperformed traditional hand-crafted features based methods~\cite{yin2012automated}. 
Effective network architecture design is the focus of these deep learning based methods.
For example, Maninis \etal~\cite{maninis2016deep} presented the DRIU network combining multi-level features to segment vessels and optic disc.
A disc-aware network \cite{fu2018disc} was designed for glaucoma screening by an ensemble of different feature streams in the network.
ResU-net was presented in \cite{shankaranarayana2017joint} with an adversarial module between the ground truth and segmentation mask to improve the final segmentation.
Based on the U-net, Fu~\etal~\cite{fu2018joint} developed the M-Net for joint OD and OC segmentation.
Although promising, CNN-based methods are often degraded when the training and testing datasets are from different domains. Our output space adversarial learning framework helps address this domain shift challenge and enhance the segmentation performance on different testing domains.

Very recently, domain adaptation techniques were explored in the field of medical image analysis~\cite{kamnitsas2017unsupervised,dou2018unsupervised,javanmardi2018domain,zhang2018task,yang2018generalizing}.
Previous methods~\cite{kamnitsas2017unsupervised, dou2018unsupervised} performed the latent feature alignment to explore a shared feature space on the source and target domain through the adversarial learning.
Another cut-in point for domain adaptation is to transfer the images from the target domain to the source domain, and then to apply the trained network to the transferred images~\cite{zhang2018task, zhang2018translating, chen2018semantic}.
Among these methods, Cycle-GAN~\cite{CycleGAN2017} is a popular technique to transfer images over different domains.
The key characteristic of these approaches is to generate style-realistic images in another domain without using paired data. Extra constraints are needed to guide this unsupervised style transfer process.
For example, Zhang~\etal~\cite{zhang2018translating} employed two segmentation networks stacked behind the cycle-GAN to act as an extra supervision on the generators to enhance the shape-consistency.
In~\cite{chen2018semantic}, a semantic-aware adversarial learning was introduced to prevent the semantic distortion during the image transformation.
In~\cite{zhang2018task}, a task-driven generative adversarial network was developed to enforce the segmentation consistency.
However, these methods ignore the property that for segmentation tasks, the label space (output space) of different domains are usually highly correlated in terms of the spatial structures and geometry.
Therefore, instead of exploring a shared feature space or transferring the input images, we use a patch-based output space adversarial learning to conduct the domain adaptation for joint OD and OC segmentation. 

\section{Methodology}
\label{sec:method}

Fig.~\ref{fig:overview} overviews the \textit{p}OSAL framework for joint OD and OC segmentation from retinal fundus images; our framework has three modules: an ROI extraction network $E$, a segmentation network $S$, and a patch-level discriminator $D$.
Due to the small area ratio of OD over the whole image, the ROI regions, $I_{cS}$ and $I_{cT}$, are firstly extracted from the source domain images $I_S$ and target domain images $I_T$, respectively (Section~\ref{sec:roiextraction}).
Then, the cropped source and target images $I_{cS}$ and $I_{cT}$ are fed into the segmentation network $S$ to produce the OD and OC predictions (Section~\ref{sec:segmentationnetwork}).
A patch-level discriminator $D$ is utilized to encourage the segmentation network to produce similar outputs for the source domain images $I_{cS}$ and target domain images $I_{cT}$ (Section~\ref{sec:adversariallearning}). 
The whole framework is finally optimized by adversarial learning.

\subsection{ROI Extraction}
\label{sec:roiextraction}
To perform accurate segmentation, we first locate the position of OD and then crop the disc region from the original image for further segmentation.
To achieve that, we build an extraction network $E$ to segment the OD and crop the ROI image according to the segmentation result.
The extraction network is configured to segment the optic disc to provide a rough guidance. Although only trained with the source domain images and labels, as our experiments will demonstrate, the trained extraction network generalizes well on the target domain images due to the strong and visible structure characteristics of the optic disc in both source and target domain images. 
Therefore, the disc regions of both domain images can be obtained by the same extraction network.
Specifically, our extraction network follows a U-Net~\cite{ronneberger2015u} architecture and is trained with resized source images ($640\times640$) and corresponding OD labels. The trained U-net can be used for coarse OD prediction in both domains.
We then map the predicted OD mask back to the original image and crop a sub-image with the size of $512\times512$ based on the center of the predicted OD mask.
The extraction network $E$ has $19$ convolutional layers, and the last one is a $1\times1$ convolutional layer with one output feature channel for the OD segmentation.
We use the $Sigmoid$ activation function to generate the probability map of OD.

\begin{figure}
	\centering
	\includegraphics[width=0.45\textwidth]{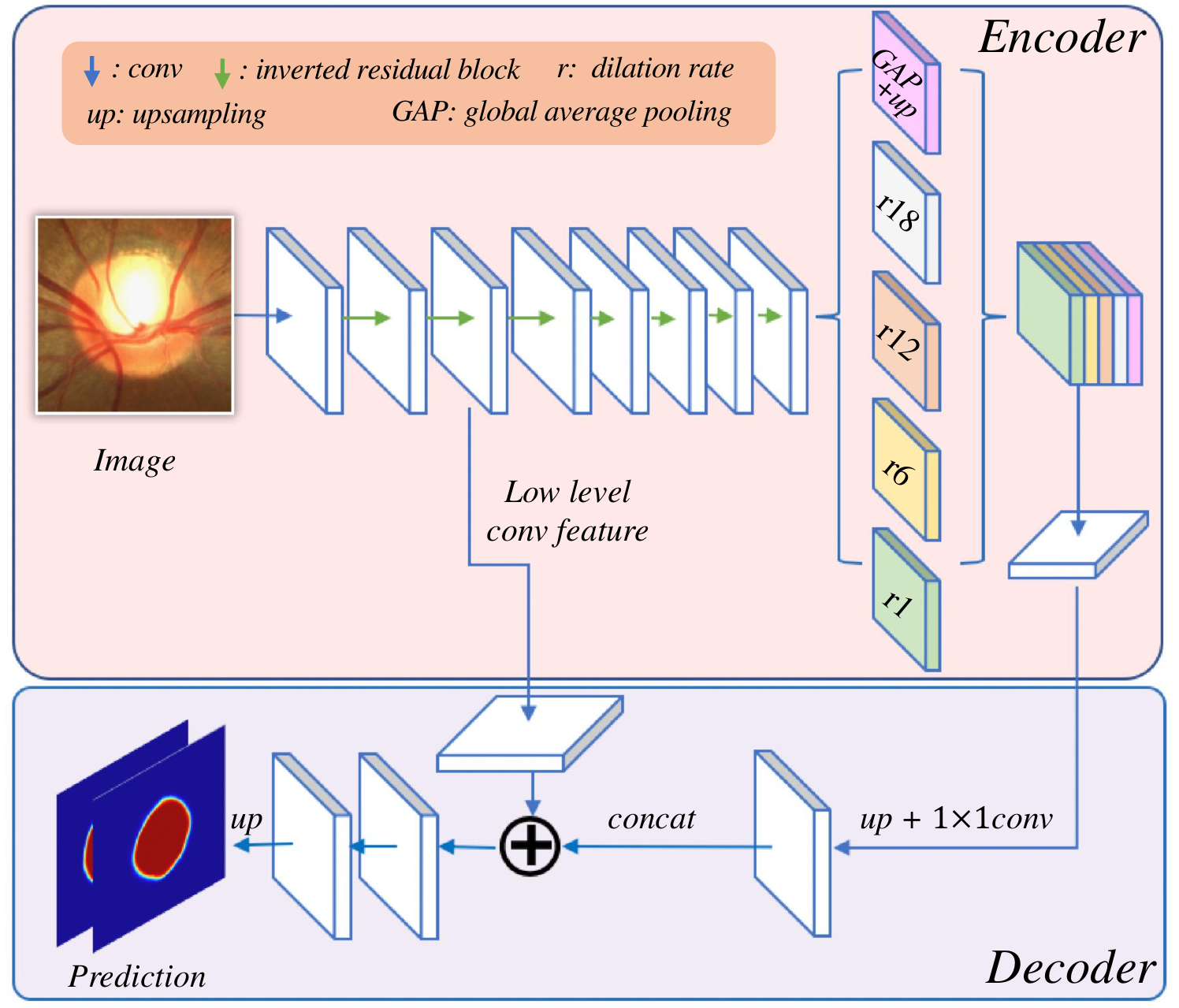}
	\caption{Architecture of the segmentation network. It is based on DeepLabv3+ but with MobileNetV2 as the network backbone. 
	} 
	\label{fig:seg_network}
	\vspace{-0.3cm}
\end{figure}

\subsection{Segmentation Network with Morphology-aware  Loss}
\label{sec:segmentationnetwork}

We conduct the OD and OC segmentation based on the above-cropped ROI images.
To better capture the geometric structure of the output space, we customize a network with a novel morphology-aware segmentation loss for high-quality segmentation of the OD and OC.

\subsubsection{Segmentation Network Architecture} 
our segmentation network follows the spirit of DeepLabv3+ architecture~\cite{chen2018encoder}. 
To further reduce the number of parameters and the computation cost, we replace the backbone network Xception~\cite{chen2018encoder} with the lightweight and handy MobileNetV2~\cite{sandler2018mobilenetv2} as shown in Fig.~\ref{fig:seg_network}.
The first initial convolutional layer and the following seven inverted residual blocks of the MobileNetV2 are utilized to extract features. We keep the stride of the first convolutional layer and the following three blocks as the initial setting and set the stride as one in the remaining blocks. The total downsampling rate of the network is eight.
The ASPP component~\cite{chen2018encoder} with different dilation rates is utilized to generate multi-scale features. The feature maps are then concatenated and followed by a $1\times1$ convolutional layer.
To integrate the semantic clues from different levels, we upsample the above-combined feature and concatenate it with the low-level feature for fine-grained semantic segmentation as the DeepLabv3+ does.
Finally, we use another $1\times 1$ convolutional layer with two output channels followed by the $Sigmoid$ activation function to generate the probability maps of OD ($p^d$) and OC ($p^c$) simultaneously, according to the multi-label setting in~\cite{fu2018joint}.
The input size of the designed segmentation network is $512 \times 512 \times 3$, so that it can take the whole cropped images as input.

\subsubsection{Morphology-aware Segmentation  Loss}
to improve the segmentation, we develop a novel morphology-aware segmentation loss to guide the network to segment and capture the smoothness priors of the OD and OC.
This joint morphological loss includes a dice coefficient loss $L_{DL}$ and a smoothness loss $L_{SL}$.

The dice coefficient loss~\cite{milletari2016v} measures the overlap between the prediction and ground truth, and is written as
\begin{equation}
L_{DL}(p,y)= 1 - \frac{2 \sum _{i \in \Omega}{p_i \cdot y_i}}{\sum _{i \in \Omega} p_i^2 + \sum _{i \in \Omega} y_i^2}, 
\end{equation}
where $\Omega$ are the total pixels in the image; $p$ and $y$ are the predicted probability map and binary ground truth mask, respectively.

The smoothness loss encourages the network to produce homogeneous predictions within neighbor regions.
It is calculated by a binary pairwise label interaction: 
\begin{align}
\label{SL}
L_{SL}(p, y)& =\sum _{i \in \Omega }{\sum_{j \in \mathcal{N}^i}{B_{i,j} \times y_i \times \left | p_i - p_j \right |}},  \\ 
B_{i,j} & =\left\{\begin{matrix}
1 & if \ \ y_i = y_j \\
0 & otherwise
\end{matrix}\right. , \nonumber
\end{align}
where $\mathcal{N}^i$ is the four-connected neighbors of pixel $i$; $p$ and $y$ denote the prediction and ground truth, respectively.
The smoothness loss encourages the neighboring pixels $j$ of central pixel $i$ to have similar predicted probabilities when their ground truth belong to the same class ($B_{i,j}=1$).
Smoothness loss is applied to the OD and OC probability maps, respectively. 

The joint morphology-aware segmentation loss is defined as
\begin{align} \label{eq:L_seg}
L_{seg} & = \lambda_1 L_{DL}(p^d, y^d) + \lambda_2 L_{DL}(p^c, y^c) \\
&+ \lambda_3 [L_{SL}(p^d, y^d) + L_{SL}(p^c, y^c)], \nonumber
\end{align}
where $p^d$, $p^c$, $y^d$, $y^c$ are the predicted probability map and binary ground truth mask of OD and OC, respectively\textcolor{blue}{;}
$\lambda_1, \lambda_2, \lambda_3$ are the weights empirically set as 0.4, 0.6\textcolor{blue}{,} and 1.0, respectively. 
Observing that it is more difficult to segment OC than OD due to the unclear boundaries of OC, we thus empirically set a slight larger value for $\lambda_2$ than $\lambda_1$.

\subsection{Patch-based Output Space Adversarial Learning}
\label{sec:adversariallearning}

Different from high-level feature-based image classification, the feature for segmentation needs to encode both the low-level descriptors and high-level abstracts, such as appearance, shape, context and object semantic information. 
However, domain adaption based on feature space may not be the best choice for our segmentation task due to the complexity in handling the high-dimensional features~\cite{tsai2018learning}.
Although the image appearance shifts across domains, the segmentation of source and target domain images have similar geometry structures in the output space (\ie, segmentation mask). Therefore, bridging two domains by forcing them to share the same distribution in the output space becomes an effective way for domain adaptation. In this work, we propose to perform domain adaptation for segmentation task through the output space adversarial learning.
Specifically, the segmentation masks of target domain images should be similar to the ones of source domain.
To achieve this, we attach a patch-level discriminator $D$ after the outputs of the segmentation network $S$, and then employ the adversarial learning technique to train the whole framework.
In this adversarial setting, the segmentation network $S$ aims to fool the discriminator $D$ by generating a similar output space distribution either for source and target domain, while the discriminator aims to identify the segmentation from target domain as outliers.
The geometry structure constraints on the segmentation masks are guaranteed through this adversarial process.
\begin{figure}
	\includegraphics[width=0.45\textwidth]{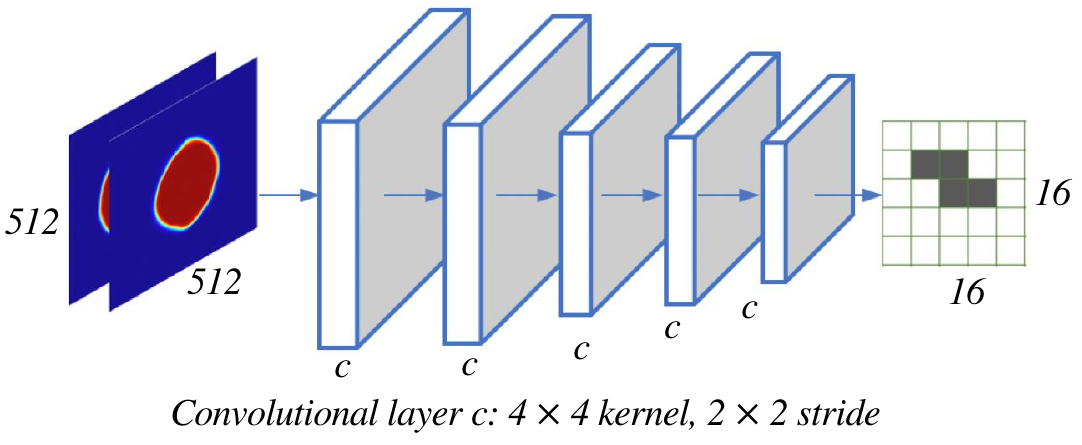}
	\centering
	\caption{Network architecture of the patch-based discriminator. 
	}
	\label{fig:adversarial_learning}
	\vspace{-0.3cm}
\end{figure}

\subsubsection{Patch Discriminator}
we employ a patch discriminator (PatchGAN)~\cite{isola2017image,li2016precomputed} to conduct the adversarial learning.
PatchGAN tries to classify whether each $m\times n$ overlapped patch from the predicted mask is in line with the distribution of that from the source predictions.
Compared with the image-level (ImageGAN) or pixel-level (PixelGAN) adversarial learning, PatchGAN has the ability to capture the local statistics~\cite{yi2017dualgan} of the output space and guides the segmentation network to focus on the local structure similarity in the image patches.

We realize the patch-based discriminator through a fully convolutional network, as shown in Fig.~\ref{fig:adversarial_learning}.
The network contains five convolutional layers with a kernel size of $4 \times 4$ and a stride of $2 \times 2$. The channel number of the five convolutional layers are $64, 128, 256, 512, 1$, respectively.
The activation function following each convolutional layer is LeakyReLU with an alpha value of $0.2$, except for the last one using the $Sigmoid$ function.
The output size ($m\times n$) of the patch-based discriminator is $16 \times 16$, in which one pixel corresponds to a patch of size $94 \times 94$ in the input probability maps.
Each patch is classified into real (1) or fake (0) through the discriminator.
We employ this adversarial learning strategy to force each generated patch in the prediction of target domain to be similar to the patch of source domain.

\subsubsection{Objective Function}
with the adversarial learning, we model the optimization as a two-player min-max game to alternately update weights in the segmentation network $S$ and the discriminator $D$.

The discriminator evaluates whether the input is from the source domain prediction. 
We formulate the training objective for the discriminator as
\begin{equation}
\label{equ:discriminator}
L_{D} =- \sum_{m,n} z \mathrm{log}(D(S(I_{cS})))+(1-z) \mathrm{log}(1-D(S(I_{cT}))),
\end{equation}
where $z = 1$ if the patch prediction is from the source domain, and $z = 0$ if the patch prediction is from the target domain. 

As for the segmentation network, the objective function consists of the proposed morphology-aware segmentation loss for the source domain images and the adversarial loss for the target domain images.
In general, the training objective of segmentation network is
\begin{align}
\label{equ:segmentation}
&L_{S} =L_{seg}(I_{cS})+L_{adv}(I_{cT}),\\
&L_{adv}(I_{cT}) = -\sum_{m,n}\mathrm{log}(D(S(I_{cT}))).  \nonumber
\end{align}
Since we have the annotations for the images from source domain, we can use the joint morphology-aware segmentation loss $L_{seg}$ to optimize the segmentation network.
The adversarial loss $L_{adv}$ is designed for the images in target domain $I_{cT}$ without any annotations.
The segmentation network is responsible for `fooling' the discriminator $D$ to classify the prediction of target domain images as the source prediction.

\subsubsection{Training Strategy}
we optimize the segmentation network and the discriminator following the standard approach from~\cite{goodfellow2014generative}.
In each training iteration, we feed the images from source domain $I_{cS}$ and target domain $I_{cT}$ to the network alternatively.
Then we optimize the whole framework by minimizing the proposed objective functions $L_S$ and $L_D$.
We repeat the above procedure for each training iteration.

\begin{table*} [!htbp]
	\centering
	\caption{Statistics of the datasets used in evaluating the proposed method.}
	\label{tab:datasetstatistics}
	\begin{tabular}{p{1.3cm}<{\centering}|p{3.5cm}<{\centering}|p{2.5cm}<{\centering}|p{2cm}<{\centering}|p{3cm}<{\centering}|p{2cm}<{\centering}}
		\hline
		Domain & Dataset &  Number of samples & Image size & Cameras & Release year  \T\B  \\
		\hline
		\T
		Source & REFUGE Train &  400 & $2124 \times 2056$  & Zeiss Visucam 500 & 2018 \B\\
		\hline
		Target & Drishti-GS Train/Test &  50 + 51 & $2047 \times 1759$ & unknown &$2014$  \T\B\\
		Target & RIM-ONE-r3 Train/Test &99 + 60& $2144 \times 1424$ & unknown & $2015$ \B\\
		Target & REFUGE Validation/Test &  400 + 400 & $1634 \times 1634$  & Canon CR-2 & $2018$ \B\\
		\hline
	\end{tabular}
\end{table*}

\begin{table*} [!htbp]
	\centering
	\caption{Results of joint OD and OC segmentation on the Drishti-GS and RIM-ONE-r3 testing datasets. 
	}
	\label{table:results}
	\begin{tabular}{p{2.8cm}<{\centering}|p{0.8cm}<{\centering}|p{0.8cm}<{\centering}|p{0.8cm}<{\centering}|p{2.8cm}<{\centering}|p{0.8cm}<{\centering}|p{0.8cm}<{\centering}|p{0.8cm}<{\centering}}
		\hline
		\multicolumn{4}{c|}{Drishti-GS} & \multicolumn{4}{c}{RIM-ONE-r3}\T\B \\
		\hline
		Method &  $DI_{cup}$  & $DI_{disc}$  & $\delta$ &  Method &  $DI_{cup}$  & $DI_{disc}$ & $\delta$ \T\B  \\
		\hline
		\textit{p}OSAL &  \textbf{0.858} & \textbf{0.965} & \textbf{0.082} &
		\textit{p}OSAL &\textbf{ 0.787} &  \textbf{0.865} & \textbf{0.081} \B\T\\   
		\textit{p}OSALseg-S  &  0.836 & 0.944 & 0.118  & 
		\textit{p}OSALseg-S  & 0.744 &  0.779 & 0.103 \B\\
		\hline
		Edupuganti \etal \cite{edupuganti2018automatic} &  0.897 & 0.967  & - &
		DRIU  \cite{maninis2016deep} &  - & 0.955 & -  \T\B\\ 
		Sevastopolsky  \cite{sevastopolsky2017optic} &  0.850 & -  & - &
		Sevastopolsky \cite{sevastopolsky2017optic} &  - & 0.950 & -  \B\\
		%
		
		%
		Son \etal \cite{son2018towards} &  - & 0.967 &- &
		Son \etal \cite{son2018towards}  &  - & 0.955 & -  \B\\

		Zilly \etal \cite{zilly2017glaucoma} &  0.871 & 0.973 & - &
		Zilly \etal \cite{zilly2017glaucoma} &  0.824 & 0.942 & - \B\\
		%
		%
		\textit{p}OSALseg-T &  \textbf{0.901}& \textbf{0.974} & 0.048 &
		\textit{p}OSALseg-T &  \textbf{0.856}& \textbf{0.968} & 0.049 \B\\

		\hline
	\end{tabular}
\end{table*}

\section{Experiments}
\label{sec:experiment}

\subsection{Dataset}

\begin{figure}[!htbp]
	\centering
	\includegraphics[width=0.45\textwidth]{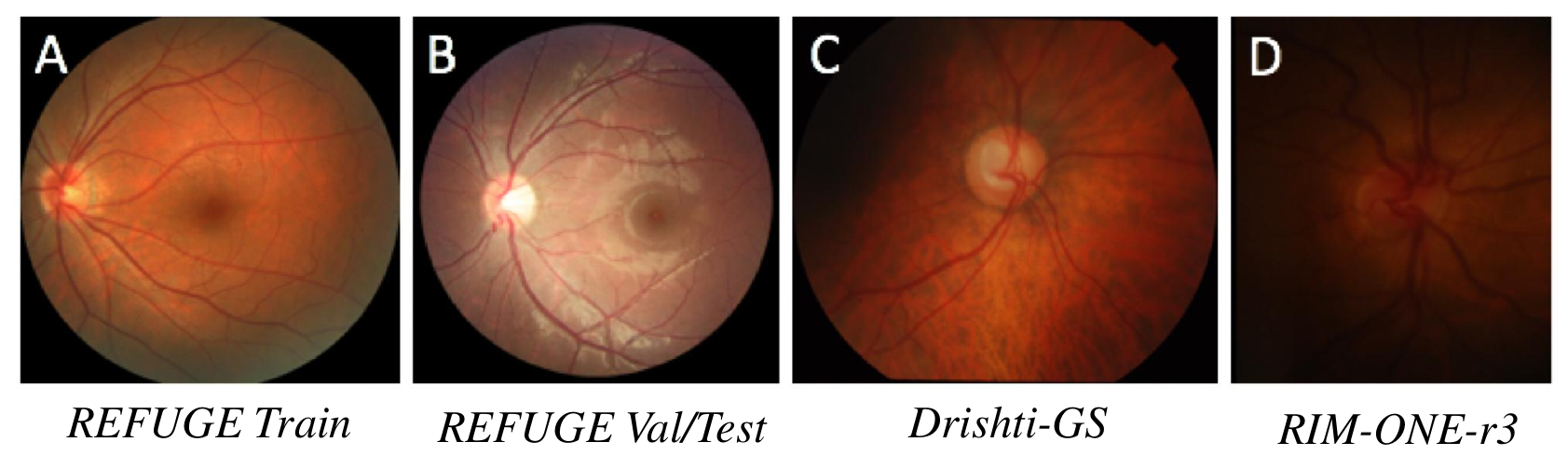}
	\caption{Comparison of images from different datasets. There exists a large variation in color and texture among the different dataset images.} 
	\label{fig:diff_input}
	\vspace{-0.5cm}
\end{figure}

We conducted experiments on three public OD and OC segmentation datasets, Drishti-GS dataset~\cite{Drishti-GS}, RIM-ONE-r3 dataset~\cite{RIM-ONE} and the REFUGE challenge dataset\footnote{\url{https://refuge.grand-challenge.org/Home/}}.
The statistics of these three datasets are listed in Table~\ref{tab:datasetstatistics}.
We refer the train part of the REFUGE dataset as the source domain, the Drishti-GS dataset, RIM-ONE-r3 dataset and the validation/test parts of the REFUGE dataset as the target domain.
The source and target domain images are captured by different cameras so that the color and texture of the images are different, as shown in Fig.~\ref{fig:diff_input}.
We first extensively evaluated and analyzed our \textit{p}OSAL framework on the Drishti-GS and RIM-ONE-r3 datasets, and then compared with other state-of-the-art segmentation methods on the REFUGE test dataset.

\subsection{Implementation Details}

The framework was implemented in Python based on Keras~\cite{chollet2015keras} with the Tensorflow backend.
We first trained the segmentation network with source domain images and annotations and then utilized the adversarial learning to train the whole \textit{p}OSAL framework in an end-to-end manner.
When training the segmentation network, we used the Adam~\cite{kingma2014adam} optimizer and initialized the backbone network weights by the MobileNetV2~\cite{sandler2018mobilenetv2} weights trained on the ImageNet dataset.
We set the initial learning rate as $1e-3$ and divided it by $0.2$ every 100 epochs. We totally trained 200 epochs with a mini-batch size of 16 on a server with four Nvidia Titan Xp GPUs.
Data augmentation was adopted to expand the training dataset by random scale, rotation, flip, elastic transformation, contrast adjustment, adding noise and random erasing \cite{zhong2017random}.
When end-to-end training the whole \textit{p}OSAL framework, we fed source and target images to the network alternatively.
The optimization method of segmentation network was the same as the above, while the discriminator $D$  was optimized with the stochastic gradient descent (SGD) algorithm.
The initial learning rate of the segmentation network and discriminator were $2.5e-5$ and $1e-5$, respectively, and decreased using the polynomial decay with a power of 0.9 as mentioned in \cite{chen2018deeplab} in a total of 100 epochs.
We conducted the morphological operation, \ie, filling the hole, to post-process the predicted mask.
The implementation and segmentation results are available in \url{https://emmaw8.github.io/pOSAL}.

\subsection{Evaluation Metrics}
We adopt the REFUGE challenge evaluation metrics, dice coefficients ($DI$) and the vertical cup to disc ratio ($CDR$), to evaluate the segmentation performance of the presented method. 
The criteria are defined as 
\begin{gather}
DI  = \frac{2 \times N_{TP}}{{2 \times N_{TP} + N_{FP} + N_{FN}}},  \\ \nonumber \\
\delta  = |CDR_p - CDR_g|,  \ \ \ \ \ \  \label{delta}
CDR  = \frac{VD_{cup}}{VD_{disc}}, 
\end{gather}
where $N_{TP}$ , $N_{FP} $\textcolor{blue}{,} and $ N_{FN}$ represent the number of true positive, false positive, and false negative pixels, respectively.
$CDR_p$ and $CDR_g$ denote the cup to disc ratio value for the prediction and ground truth, while $VD_{cup}$ and $VD_{disc}$ are the vertical diameters for OC and OD, respectively.
The dice coefficient is a standard evaluation metric for segmentation tasks, while the CDR value is one of the critical indicators for glaucoma screening in the clinical convention.
We use absolute error $\delta$ to evaluate the difference between the CDR value of prediction $CDR_p$ and that of the ground truth $CDR_g$, while the lower $\delta$ value represents the better prediction result.

\begin{figure*}[!h]
	\centering
	\includegraphics[width=0.95\textwidth]{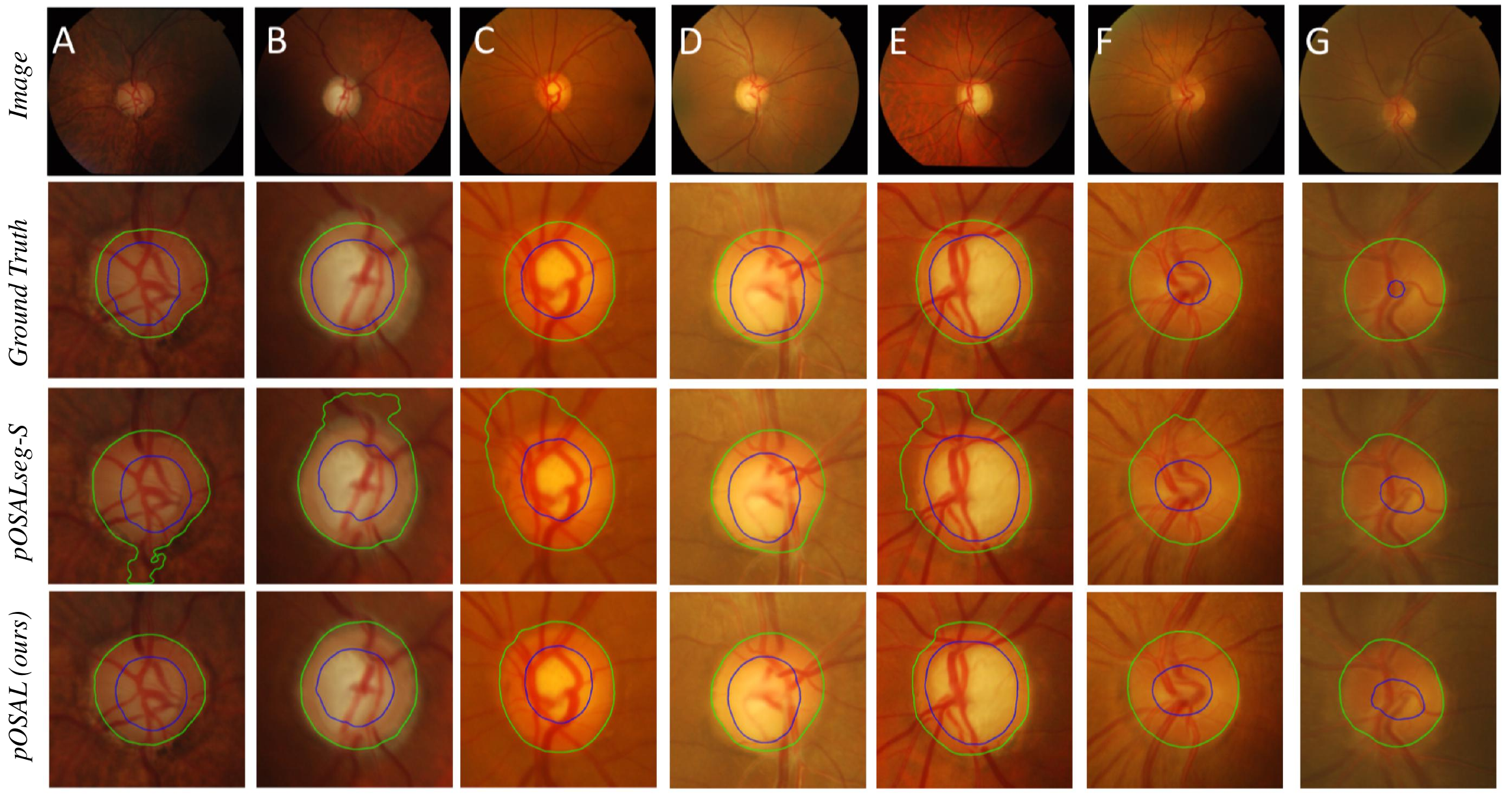}
	\caption{Qualitative results on the Drishti-GS testing dataset. Each column presents one example. From top to bottom: original image, ROI region with ground truth contours of OD and OC, results of the \textit{\textit{p}OSALseg-S}, and results of our \textit{p}OSAL framework. The green and blue contours indicate the boundary of OD and OC, respectively.}
	\label{fig:DGS-results}
\end{figure*}

\begin{table*} [!htbp]
	\centering
	\caption{Comparison with different domain adaptation methods on the Drishti-GS and RIM-ONE-r3 datasets.}
	\label{table:domainadpation}
	\begin{tabular}{p{3.5cm}<{\centering}|p{1.5cm}<{\centering}|p{1.5cm}<{\centering}|p{1.5cm}<{\centering}|p{1.5cm}<{\centering}|p{1.5cm}<{\centering}|p{1.5cm}<{\centering}}
		\hline
		\multirow{2}{*}{Method} & \multicolumn{3}{c|}{Drishti-GS} & \multicolumn{3}{c}{RIM-ONE-r3}\T\B \\
		\cline{2-7} & $DI_{cup}$  & $DI_{disc}$ & $\delta$  &  $DI_{cup}$  & $DI_{disc}$ & $\delta$  \T\B  \\
		\hline
		TD-GAN\cite{zhang2018task} &  0.747 & 0.924&0.117&0.728&0.853     &0.118 \T\B\\
		Hoffman~\etal~\cite{hoffman2016fcns}&0.851&0.959&0.093&0.755&0.852    &0.082 \B\\
		Javanmardi~\etal~\cite{javanmardi2018domain} &  0.849 & 0.961 &0.091&0.779&0.853    &0.085 \B\\
		OSAL-pixel &  0.851 & 0.962 &0.089 & 0.778&0.854    &0.084  \B\\ \hline
		\textit{p}OSAL (ours) &  \textbf{0.858} & \textbf{0.965} & \textbf{0.082} &  \textbf{0.787} &\textbf{0.865} & \textbf{0.081}    \T\B\\
		\hline
	\end{tabular}
\end{table*}

\subsection{Experiments on Drishti-GS and RIM-ONE-r3 Datasets}
Under the domain adaptation setting, we need to utilize the unlabeled target domain images to train the whole framework.
For a fair comparison, the unlabeled target domain images used in the training phase were different from the target domain images in the testing phase. 
We follow this experiment setting over our experiments. 

\subsubsection{Effectiveness of Patch-based Output Space Adversarial Learning}
the Drishti-GS and RIM-ONE-r3 datasets both provide the training and testing images splits. 
Therefore, for the Drishti-GS dataset, we used the REFUGE training dataset as the source domain and the training part of the Drishti-GS dataset as the target domain to train our \textit{p}OSAL framework. 
We then report the segmentation performance of our method on the testing dataset of Drishti-GS.
We conducted experiments with the same dataset setting for the RIM-ONE-r3 dataset.

Table~\ref{table:results} presents the segmentation results on the Drishti-GS and RIM-ONE-r3 testing datasets.
For each dataset, we show the segmentation performance of our \textit{p}OSAL framework (\textit{p}OSAL) and the segmentation network only (\textit{\textit{p}OSALseg-S}) to demonstrate the effect of the proposed output space adversarial learning. 
%
We used the REFUGE training dataset to train the \textit{\textit{p}OSALseg-S} model and directly evaluated it on the Drishti-GS and RIM-ONE-r3 testing images. 
It is observed that the \textit{p}OSAL consistently improves the DI of optic cup and disc and the $\delta$ on the Drishti-GS and RIM-ONE-r3 datasets compared with \textit{\textit{p}OSALseg-S}.
On the RIM-ONE-r3 dataset, we achieve 4.3\% and 8.6\% DI improvement for the cup and disc segmentation with the patch-based output space adversarial learning, while we also achieve 2.2\% and 2.1\% DI improvement for OC and OD on the Drishti-GS dataset, respectively.
Since the domain discrepancy between the REFUGE training data and RIM-ONE-r3 data is larger than the difference between REFUGE training data and Drishti-GS data (see Fig.~\ref{fig:diff_input}), the absolute DI values of optic cup and disc on RIM-ONE-r3 is lower than those on Drishti-GS.
%
These comparisons demonstrate the patch-based output space adversarial learning can alleviate the performance degradation among the datasets with domain shift.

\subsubsection{Qualitative Results}
we show some qualitative results of the optic OD and OC segmentation on the Drishti-GS dataset in Fig.~\ref{fig:DGS-results}.
For the \textit{\textit{p}OSALseg-S} method without domain adaptation, it can locate the approximate location but fails to generate accurate boundaries of OD and OC due to the low image contrast at the boundary between OD and OC, as well as between OD and background (especially columns A, B and E in Fig.~\ref{fig:DGS-results}).
In contrast, our proposed method successfully localizes the OD and OC and further preserves the shape prior and generates more accurate boundaries.
%

\begin{table*} [!htbp]
	\centering
	\caption{Results of OD and OC segmentation on the REFUGE testing dataset. Top three items are bold for each metric.}
	\label{tab:results-REFUGE}
	\begin{tabular}{c|cccccc|cc}
		\hline
		\small
		Team &$DI_{cup}$  &$R_{DI_{cup}}$ &$DI_{disc}$ &$R_{DI_{disc}}$ &$\delta$ &$R_{\delta}$ &$S_f$ &Overall Rank \T\B  \\
		\hline
		CUHKMED (ours)&\textbf{0.8826} & 2 &\textbf{0.9602}& 1 &\textbf{0.0450}    &2 &\textbf{1.75} &1    \T\B\\
		\hline
		Masker    &\textbf{0.8837}    & 1 &0.9464         & 7 &\textbf{0.0414}&1 &\textbf{2.50} &2 \T\B\\
		BUCT     &\textbf{0.8728}     & 3 &\textbf{0.9525}& 3 &\textbf{0.0456}&3 &\textbf{3.00} &3    \B\\
		NKSG     & 0.8643             & 5 &0.9488         & 5 &0.0465         &4 &4.60 &4    \B\\
		VRT     & 0.8600             & 6 &\textbf{0.9532}& 2 &0.0525         &7 &5.40 &5    \B\\
		AIML    & 0.8519             & 7 &0.9505         & 4 &0.0469         &5 &5.45 &6    \B\\
		Mammoth & 0.8667             & 4 &0.9361         & 10&0.0526         &8 &7.10 &7    \B\\
		SMILEDeepDR & 0.8367         & 8 &0.9386         & 9 &0.0488         &6 &7.45 &8    \B\\
		NIGHTOwl & 0.8257             & 10&0.9487         & 6 &0.0563         &9 &8.60 &9    \B\\
		SDSAIRC & 0.8315             & 9 &0.9436         & 8 &0.0674         &10&9.15 &10\B\\
		Cvblab     & 0.7728             & 11&0.9077         & 11&0.0798         &11&11.00&11\B\\
		Winter\_Fell & 0.6861         & 12&0.8772         & 12&0.1536         &12&12.00&12\B\\
		\hline
	\end{tabular}
\end{table*}

\subsubsection{Comparison with other Segmentation Methods}
we also report the segmentation performance of some supervised learning methods in literature for the above two datasets.
In these methods, the networks were trained with the training split of the dataset in a supervised way and evaluated on the testing part of the related datasets.
Besides the methods in the literature, we also trained our segmentation network with the training data and report the segmentation performance on the testing images (denoted as \textit{\textit{p}OSALseg-T}) to show the effectiveness of our designed segmentation network with the morphology-aware segmentation loss. 
We show these results in Table~\ref{table:results}. 
As we can see our segmentation network (\textit{\textit{p}OSALseg-T}) can produce better DI for the optic cup and disc segmentation compared with other supervised methods on both of the Drishti-GS and RIM-ONE-r3 datasets, showing the effectiveness of the segmentation network design.
In another aspect, it is observed that the optic cup and disc segmentation performance of our \textit{p}OSAL framework on the Drishti-GS dataset is very close to that of these supervised methods, which further indicates the effectiveness of the proposed patch-based output space adversarial learning.

\subsubsection{Comparison with Different Domain Adaptation Approaches}
%
as far as we know, we are not aware of any previous works that explores domain adaptation for optic disc and cup segmentation. Therefore, we compared our \textit{p}OSAL framework with several unsupervised domain adaptation ideas in other medical image analysis and natural image processing tasks.
Specifically, we compared our \textit{p}OSAL framework with a Cycle-GAN based unsupervised domain adaptation method TD-GAN~\cite{zhang2018task}, a latent feature alignment method~\cite{hoffman2016fcns}, and a recent domain adaptation method for eye vasculture segmentation~\cite{javanmardi2018domain}. 
To show the effectiveness of our patch-based discriminator, we also implemented a pixel-based discriminator for adversarial learning (denoted as \textit{OSAL-pixel}).
Table~\ref{table:domainadpation} presents the performance of different domain adaptation methods on the Drishti-GS and RIM-ONE-r3 datasets.
All the methods adopted the same segmentation network architecture for a fair comparison.
As we can see, our \textit{p}OSAL framework achieves the best performance for the optic cup and disc segmentation among these unsupervised domain adaptation methods on the Drishti-GS and RIM-ONE-r3 datasets. 
In another aspect, the patch-based adversarial learning outperforms the pixel-level discriminator (\textit{OSAL-pixel}) and the image-level discriminator (Javanmardi~\etal~\cite{javanmardi2018domain}) for adversarial learning, as it considers the local and global context information simultaneously.

\subsubsection{Performance of Glaucoma Screening}
\begin{figure}
	\centering
	\includegraphics[width=0.45\textwidth]{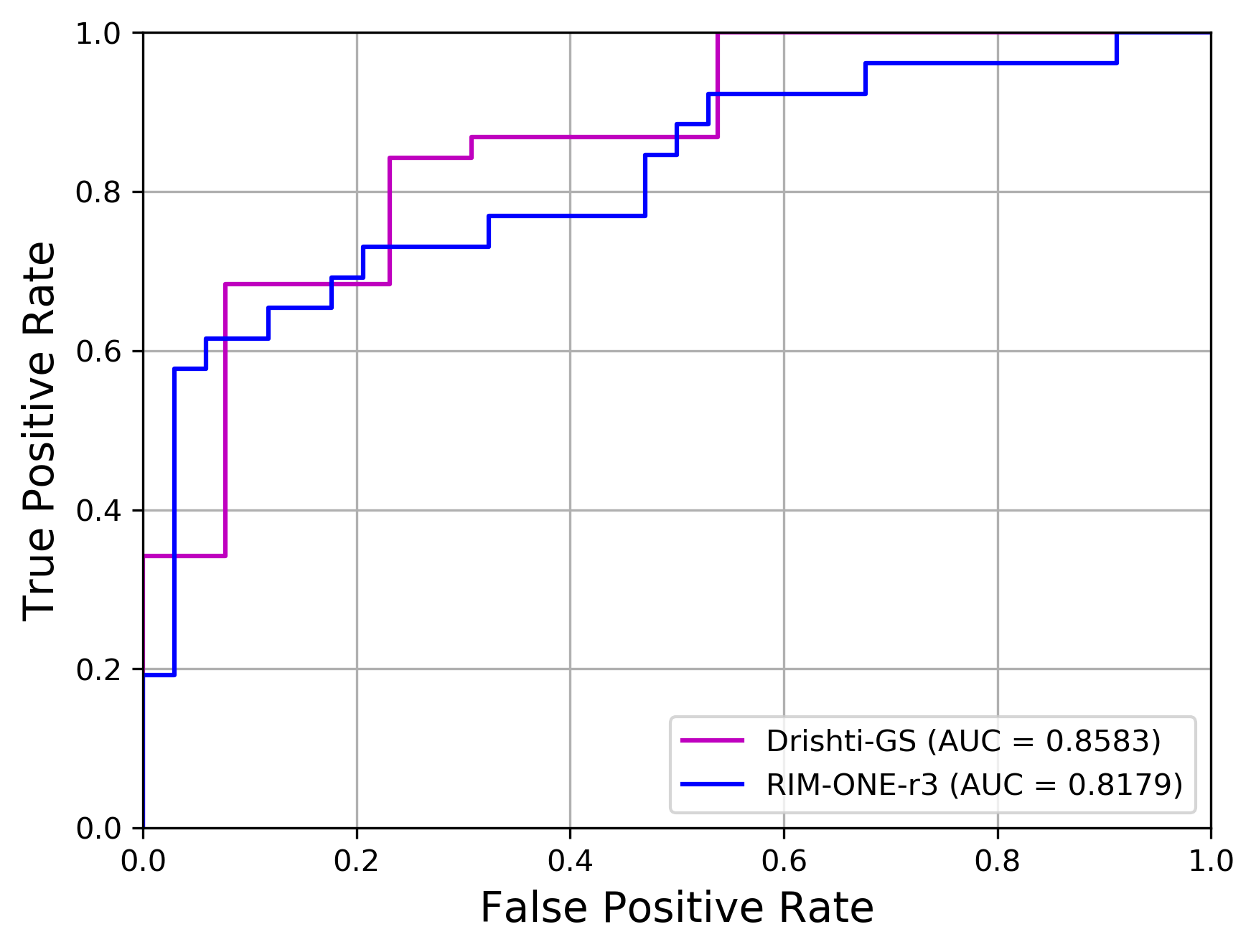}
	\centering
	\caption{The ROC curves of our method in glaucoma screening on the Drishti-GS and RIM-ONE-r3 datasets.} 
	\label{fig:ROC}
\end{figure}

the vertical CDR value is one of the important indicators for glaucoma screening. 
Therefore, we provide the glaucoma diagnose performance based on our segmentation method.
Specifically, we make use of the segmented OD and OC masks to calculate the vertical CDR value $p_i$ for the $i$th image.
Then the normalized CDR values $\hat{p}_i $ of the $i$th image can be calculated using
\begin{equation}
\hat{p}_i = \frac{p_i-p_{min}}{p_{max}-p_{min}},
\end{equation}
where $p_{max} $ and $p_{min} $ are the maximum and minimum vertical CDR values, respectively, through all the testing images.
We report the Receiver Operating Characteristic (ROC) curve and Area Under ROC Curve (AUC) for glaucoma screening evaluation in Fig.~{\ref{fig:ROC}}. 


\subsection{Results of the REFUGE Challenge}


We report the results for the optic disc and cup segmentation task of the REFUGE challenge in conjunction with MICCAI 2018.
The challenge datasets consist of three parts: a training dataset, a validation dataset, and a testing dataset.
The validation and testing datasets are acquired with the same cameras and the detailed information is shown in Table~\ref{tab:datasetstatistics}.
The testing images were evaluated via an on-site part, where the participants had four hours to acquire the testing images and submit the prediction results to avoid manually tuning the hyper-parameters. 
We treated the training images as the source domain and the validation images as the unlabeled target domain to train our \textit{p}OSAL framework.
The testing image prediction was then acquired by an ensemble of five models to improve the segmentation performance further.
Other participant teams also utilized the ensemble scheme to generate the final testing prediction (\eg, team Masker).
%

\begin{figure}
	\centering
	\includegraphics[width=0.45\textwidth]{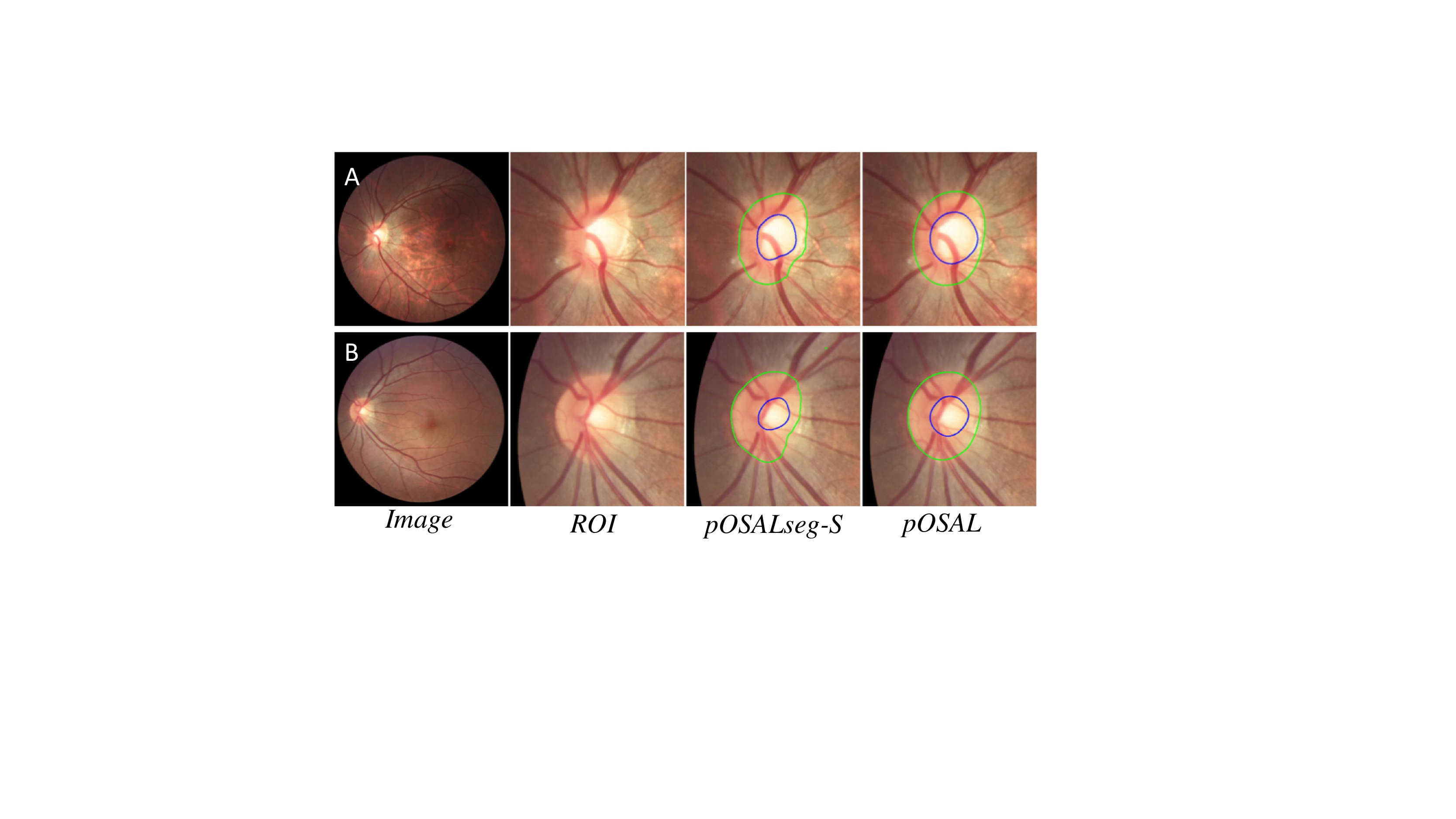}
	\caption{Qualitative results of the REFUGE testing image. The green and blue contours indicate the boundary of OD and OC, respectively.} 
	\label{fig:REFUGE-results}
\end{figure}

There were 12 teams selected to participate in the onsite REFUGE challenge for the OD and OC segmentation task, and the challenge results are listed in Table~\ref{tab:results-REFUGE}
(The leaderboard is in the challenge website\footnote{\url{https://refuge.grand-challenge.org/Results-Onsite\_TestSet/}}).
Each team was only allowed for one submission, and the teams were ranked according to the following weighted sum of three metrics:
\begin{equation}
S_f = 0.35 \times R_{DI_{cup}} + 0.25 \times R_{DI_{disc}} + 0.4 \times R_{\delta},
\end{equation}
where $R_{DI_{cup}}$, $R_{DI_{disc}}$\textcolor{blue}{,} and $R_{\delta}$ denote the rank of $DI_{cup}$,  $DI_{disc}$ and $\delta$ criteria, respectively.
A lower $S_f$ suggests a better final rank. 
All these methods utilized deep neural networks for OD and OC segmentation.
Some methods made use of other datasets (\eg, ORIGA~\cite{zhang2010origa} and IDRiD\footnote{\url{https://idrid.grand-challenge.org}}) as the extra training data to improve the model generalization capability, while we only used the training and validation datasets provided by the organizer.
In Table~\ref{tab:results-REFUGE}, it is observed that our \textit{p}OSAL framework outperforms the second-ranking team Masker by around 1.4\% on the optic disc DI, while we achieve compelling performance on both the DI of optic cup and CDR $\delta$. 
Overall, our \textit{p}OSAL framework achieves the best overall ranking score $S_f$ and the first place in this challenging task, demonstrating the effectiveness of \textit{p}OSAL.
We also visually compare the appearance difference of the results with and without the output space adversarial learning using a single model.
As shown in Fig.~{\ref{fig:REFUGE-results}}, our \textit{p}OSAL framework could retain the elliptical features and push the optic cup within the optic disc to produce better visual results.

\begin{table} [!htbp]
	\centering
	\caption{Results of OD and OC segmentation on the REFUGE validation dataset.}\label{Table:REFUGE_val_comparison}
	\begin{tabular}{p{2.5cm}<{\centering}|p{1.2cm}<{\centering}|p{1.2cm}<{\centering}|p{1.2cm}<{\centering}}
		\hline
		\small
		Method &  $DI_{cup}$  & $DI_{disc}$ & $\delta$ \T\B  \\
		\hline
		\textit{p}OSALseg-S &  0.869 & 0.932 &0.059 \T\B\\
		\textit{p}OSAL &  \textbf{0.875} & \textbf{0.946} & \textbf{0.051} \B\\
		\hline
	\end{tabular}
\end{table}

Besides, we further validate the effectiveness of the patch-based output space adversarial learning on the REFUGE validation dataset.
Specifically, we randomly divided the 400 validation images into two equal-sized parts and used them as the unlabeled target domain training data to train the network and the target domain testing data to evaluate the network, respectively.
We report the performance of our \textit{p}OSAL framework and the same network without domain adaptation (\textit{\textit{p}OSALseg-S}) in Table~\ref{Table:REFUGE_val_comparison}.
As we can see, the presented \textit{p}OSAL framework also improves the DI for optic cup and disc on the  REFUGE validation dataset.

\begin{table} [!htbp]
	\centering
	\caption{Comparison of different loss functions.}
	\label{Table:loss}
	\begin{tabular}{p{3.0cm}<{\centering}|p{1.5cm}<{\centering}|p{1.5cm}<{\centering}}
		\hline
		\small
		Loss function &  $DI_{cup}$  & $DI_{disc}$ \T\B  \\
		\hline
		Cross Entropy Loss & 0.860  & 0.953 \T\B\\
		Dice Loss  & 0.878 & 0.950  \B\\
		Morphology-aware  Loss  &\textbf{0.885} & \textbf{0.956} \B\\
		
		\hline
	\end{tabular}
\end{table}

We compared the effect of different loss functions to the segmentation network.
Specifically, we divided the 400 REFUGE training images into 320 and 80 images to train and evaluate the network, respectively, with different loss functions.
The results are shown in Table~\ref{Table:loss}.
We can find that the {Dice Loss} achieved a better DI for OC and a comparable DI for OD compared with the Cross Entropy Loss.
When combined with the smooth loss, the proposed morphology-aware segmentation loss achieves the best DI of OD and OC predictions, suggesting that the morphology-aware segmentation loss produces high-quality predictions. 

We also provide the glaucoma screening evaluation results here for readers' reference.
We directly utilized the segmentation results of our \textit{p}OSAL framework to calculate the vertical CDR values to diagnose glaucoma following the method on previous two datasets. Since we cannot access the ground truth of the glaucoma, we only report the AUC of glaucoma screening on the challenge testing dataset. The AUC value is $0.9644$, ranking third (Team CUHKMED) in the onsite challenge\footnote{\url{https://refuge.grand-challenge.org/Results-Onsite\_TestSet/}}.

\begin{table} [h]
	\centering
	\caption{Comparison of different network backbones.}\label{tabel:model_architecture_comparison}
	\begin{tabular}{p{2.0cm}<{\centering}|p{1cm}<{\centering}|p{1cm}<{\centering}|p{1cm}<{\centering}|p{1cm}<{\centering}}
		\hline
		Block type & Params &  $DI_{cup}$  & $DI_{disc}$ &  Time \T\B  \\
		\hline
		Xception & 41.3M &  0.885 & 0.953  &0.124s  \T\B \\
		MobileNetV2 & 5.8M &  0.885 & 0.956 & 0.056s  \B \\
		\hline
	\end{tabular}
	\vspace{-0.3cm}
\end{table}

\section{Discussion}
\label{sec:discussion}

The optic disc to cup ratio has been recognized as an essential attribute for glaucoma screening, so a high-quality automatic segmentation method is highly demanded in clinical practice.
Although plenty of works worked on this problem, there still exists a gap between research works and clinical practice due to the lack of annotations, the noisy or sparse annotations of clinical applications, and the domain shift between training images and real testing images.
In this work, we focus on developing unsupervised domain adaptation methods to enable optic disc and cup segmentation applied to clinical applications.
The key insight of our method is to encourage the target domain predictions closer to the source ones, since the OD and OC geometry structure should be reserved for source and target domain images.
The extensive experiments on three public fundus image datasets have sufficiently demonstrated the potential of our method in generalizing the segmentation network to unlabeled target domain images.


\begin{table} [t]
	\centering
	\caption{Performance of the Extraction Network \textit{E}.}
	\label{tabel:results_E}
	\begin{tabular}{c|c|c|c|c|c|c}
		\hline
		\multirow{2}{*}{Method} & \multicolumn{3}{c|}{Drishti-GS} & \multicolumn{3}{c}{RIM-ONE-r3}\T\B \\
		\cline{2-7} & $DI_{cup}$  & $DI_{disc}$  & $\delta$  &  $DI_{cup}$  & $DI_{disc}$ & $\delta$  \T\B  \\
		\hline
		\textit{E} &  0.798 &   0.930&0.098&0.638&0.709  & 0.150  \T\B\\
		\textit{p}OSAL &  0.858 & 0.965& 0.082&
		0.787&0.865  & 0.081 \B\\
		\hline
	\end{tabular}
\end{table}

In our method, we first used an extraction network to crop an ROI image before performing the segmentation.
To show the necessity of the ROI extraction, we conducted another experiment to see the overall performance of the extraction network $E$. 
We trained a new extraction network $E$ with two outputs instead of only the optic disc. 
The performance of OC and OD on the Drishti-GS and RIM-ONE-r3 datasets are shown in Table~\ref{tabel:results_E}.
It is observed that the segmentation performance of using only the extraction network $E$ is much lower than the full method \textit{p}OSAL.
The results here verify the effectiveness of the two-stage pipeline.
A good ROI is the basis of good segmentation results in the two-stage pipeline. In some cases, the boundary between the OD and background is unclear, so the OD may not in the center of the ROI. 
To avoid this kind of situation, the ROI size is needed to be designed properly. 
In our experiments, the width and height of ROI are about twice larger than that of the OD, which helps relax the location deviation. We found that all of the OD and OC regions are covered by the cropped ROI under this experiment setting.

Currently, numerous works are focusing on computation-efficient network design~\cite{sandler2018mobilenetv2,ma2018shufflenet} to promote the deep learning applications for mobile devices with limited computing power. 
In our work, we used a MobileNetV2~\cite{sandler2018mobilenetv2} as the network backbone to reduce the computation cost.
We compared the segmentation performance, parameter numbers, and testing time cost of the original backbone: Xception~\cite{chen2018encoder} and MobileNetV2~\cite{sandler2018mobilenetv2} in Table~\ref{tabel:model_architecture_comparison}. 
It is observed that the MobileNetV2 backbone has fewer parameters and can reduce the testing time by half with similar performance compared with the Xception backbone.
This comparison indicates that we could develop more lightweight network architecture to promote the development of mobile applications for glaucoma screening.

Although our network can be generalized to unlabeled target domain images, collecting extra unlabeled images from the target domain is needed to train the network.
Moreover, it is necessary to re-train a new network when the image comes from a new target domain.
In practice, the unlabeled target domain images may not be available during the training stage.
Therefore, in the future, we would explore the domain generalization techniques~\cite{muandet2013domain,li2017deeper,li2018domain} to tackle this problem without the demand for many target images.

\section{Conclusion}
\label{sec:conclusion}

We presented a novel patch-based Output Space Adversarial Learning framework to segment optic disc and cup from different fundus images. 
We first employed a lightweight and efficient network with the morphology-aware segmentation loss to generate accurate and smooth predictions.
To tackle the domain shift between the source and target domains, we exploited unsupervised domain adaptation model to improve the generalization of the segmentation network.
Particularly, the patch-based output space adversarial learning was designed to capture the local statistics of the output space and guide the segmentation network generate similar outputs for the images from the target and source domains. 
We also performed extensive experiments on three public retinal fundus image datasets to demonstrate the significant improvements and the effectiveness of the presented \textit{p}OSAL framework.
More effort will be involved to extend this framework to other medical image analysis problems in the near future.

\appendices

\ifCLASSOPTIONcaptionsoff
  \newpage
\fi



%
%
%

\bibliography{ref}

%




\end{document}